\documentclass[preprint,12pt]{elsarticle}

\usepackage[numbers]{natbib}

\usepackage{times}
\usepackage{epsfig}
\usepackage{amssymb}
\usepackage{booktabs}
\usepackage{tabularx}
\usepackage{multirow}
\usepackage{subcaption}
\usepackage[export]{adjustbox}
\usepackage{float}
\usepackage{dashrule}
\usepackage{arydshln}
\usepackage[utf8x]{inputenc}
\usepackage[space]{grffile}
\usepackage{anyfontsize}
\usepackage{t1enc}
\usepackage{mathtools, nccmath}

\usepackage{amsmath,amsfonts}
\usepackage{algorithmic}
\usepackage{algorithm}
\usepackage{array}
\usepackage{textcomp}
\usepackage{stfloats}
\usepackage{url}
\usepackage{verbatim}
\usepackage{graphicx}
\usepackage{animate}
\usepackage{xcolor}

\journal{Neurocomputing}

\begin{document}

\begin{frontmatter}


\title{WAIT: Feature Warping for Animation to Illustration video Translation using GANs}

\author[1]{Samet~Hicsonmez}
\ead{samethicsonmez@hacettepe.edu.tr}
\affiliation[1]{organization={Hacettepe University},
            addressline={Dept. of Computer Engineering}, 
            city={Ankara},
            country={Turkey}}

\author[2]{Nermin~Samet}
\ead{nermin@ceng.metu.edu.tr}
\affiliation[2]{organization={Middle East Technical University},
            addressline={Dept. of Computer Engineering}, 
            city={Ankara},
            country={Turkey}}

\author[1]{Fidan~Samet}
\ead{fidanlsamet@gmail.com}

\author[1]{Oguz~Bakir}
\ead{oguzbakir0@gmail.com}

\author[2]{Emre~Akbas}
\ead{emre@ceng.metu.edu.tr}

\author[1]{Pinar~Duygulu}
\ead{pinar@cs.hacettepe.edu.tr}

\begin{abstract}
In this paper, we explore a new domain for video-to-video translation. Motivated by the availability of animation movies that are adopted from illustrated books for children, we aim to stylize these videos with the style of the original illustrations. Current state-of-the-art video-to-video translation models rely on having a video sequence or a single style image to stylize an input video. We introduce a new problem for video stylizing where an unordered set of images are used. This is a challenging task for two reasons: i) we do not have the advantage of temporal consistency as in video sequences; ii) it is more difficult to obtain consistent styles for video frames from a set of unordered images compared to using a single image.   

Most of the video-to-video translation methods are built on an image-to-image translation model, and integrate additional networks such as optical flow, or temporal predictors to capture temporal relations. These additional networks make the model training and inference complicated and slow down the process. To ensure temporal coherency in video-to-video style transfer, we propose a new generator network with feature warping layers which overcomes the limitations of the previous methods. We show the effectiveness of our method on three datasets both qualitatively and quantitatively. Code and pretrained models are available at {https://github.com/giddyyupp/wait.}
\end{abstract}

% Keywords

\end{frontmatter}

\section{Introduction}

\begin{figure*}[h]
\captionsetup[subfigure]{labelformat=empty}
\centering
\setlength\tabcolsep{1.5pt} % default value: 6pt
\resizebox{1.0\textwidth}{!}{
\begin{tabular}{cc}
{Animation Videos} & {Illustrations} \\

{\rotatebox[origin=t]{90}{\textit{\textbf{\small AS}}}} \includegraphics[width=0.60\textwidth,  ,valign=m, keepaspectratio,] {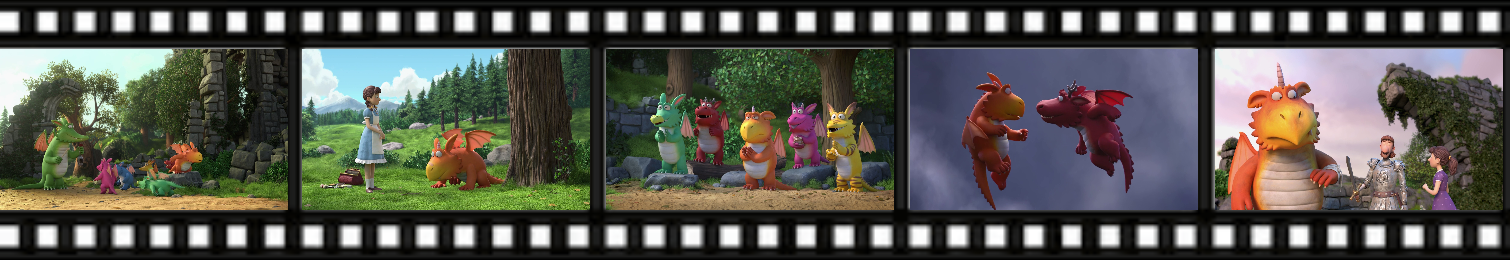} & 
\multirow[t]{2}{*}[16pt]{\includegraphics[width=0.40\textwidth,  ,valign=t, keepaspectratio,] {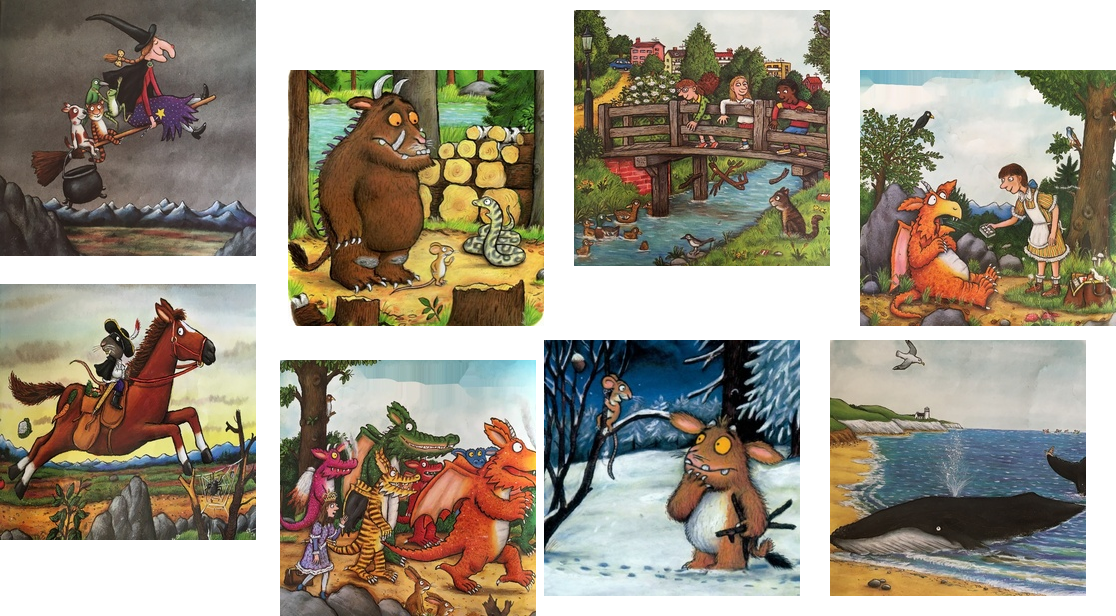}}      \\
{\rotatebox[origin=t]{90}{\textit{\textbf{\small AS}}}} \includegraphics[width=0.60\textwidth,  ,valign=m, keepaspectratio,] {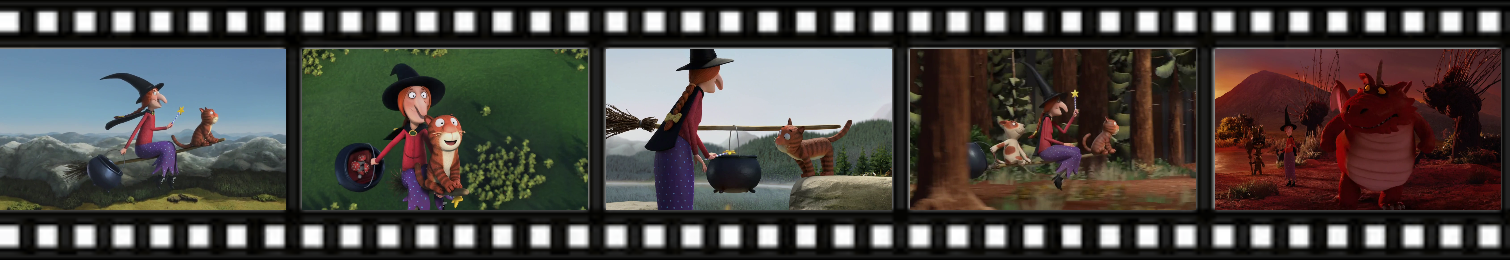} &  
\\ \\
% \hline
{\rotatebox[origin=t]{90}{\textit{\textbf{\small BP}}}} \includegraphics[width=0.60\textwidth,  ,valign=m, keepaspectratio,] {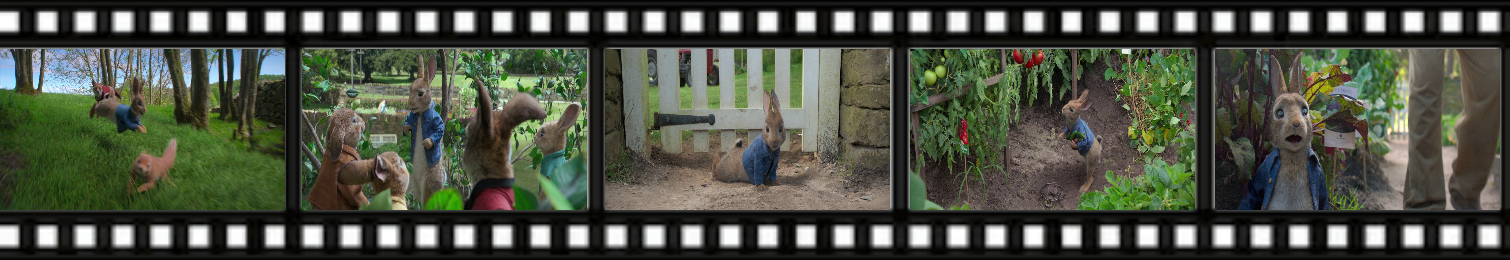} & 
\multirow[t]{2}{*}[17pt]{\includegraphics[width=0.40\textwidth,  ,valign=t, keepaspectratio,] {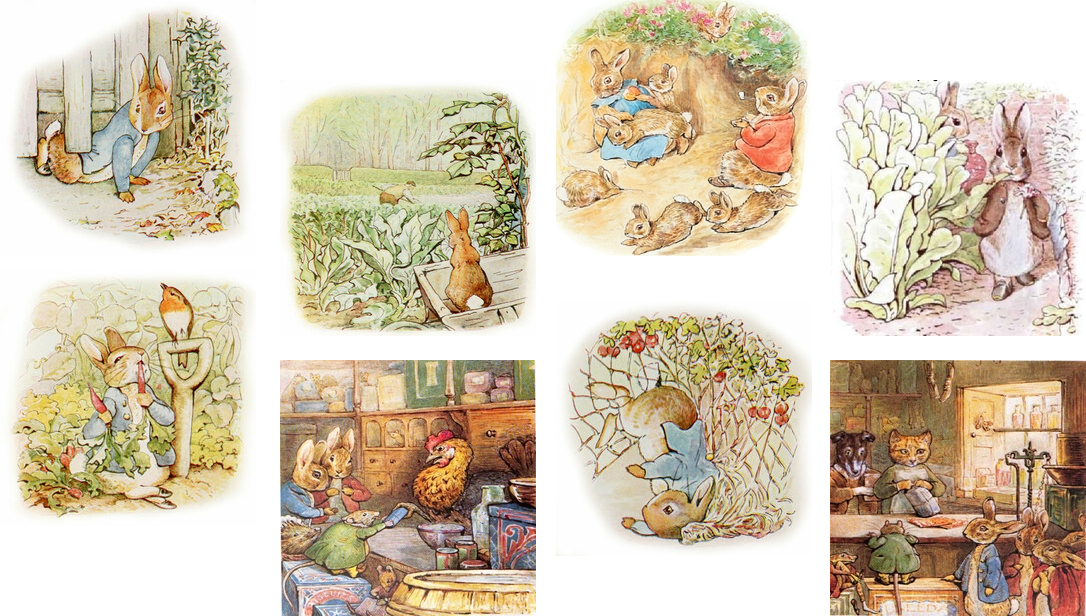}}     \\
{\rotatebox[origin=t]{90}{\textit{\textbf{\small BP}}}} \includegraphics[width=0.60\textwidth,  ,valign=m, keepaspectratio,] {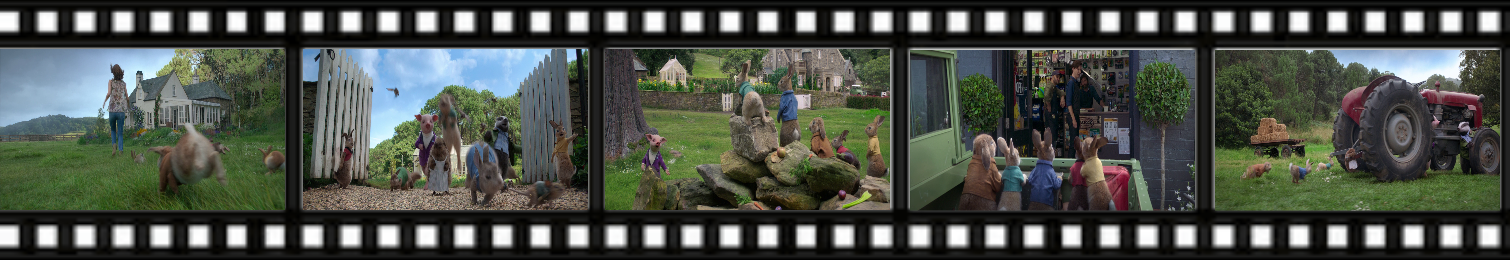} &  \\
\end{tabular}}
\caption{Left: Sequences from animation movies adopted from stories illustrated by Axel Scheffler (AS) and Beatrix Potter (BP) where temporal relation exists. Right: Pages taken from various books corresponding to the same illustrators. Note that, even when the same characters are present both in animations and illustrations, their resemblance is limited. Colors and styles are also different in movies and books. Moreover, on BP dataset, in some illustrations corners are left blank. }
\label{fig:sample-ill}
\end{figure*}

Recently, some of the children's books have been adopted as animated movies. ``Peter Rabbit" of Beatrix Potter and ``ZOG" of Axel Scheffler are two of the examples (see Figure~\ref{fig:sample-ill}). Although the characters are the same, the styles of the illustrations in books are different from animations. 
Unfortunately, for those who grow up with the original books, the new adaptations don't give the same taste as they are not familiar. 
In order to generate videos for new content, with the taste of the original style, we explore the problem of converting animation movies adopted from children books into stylized versions of their original illustrations.

In this study, we address this challenge by exploring the use of unordered set of images in the target domain as an alternative to a single image or a sequence of images from a target video.
This setting, although being more difficult, overcomes two major problems faced by other approaches: (i) the need for collecting sequential data that is not always available and costly, and (ii) the difficulty of extracting style information using just a single image.  

%%%%%%%%%%% Figure DATASET RESIMLERI %%%%%%%%

\newpage
The problem of converting a given video in the source domain to a stylized version in the target domain can be attacked with the style transfer approaches. A trivial solution would involve using image-to-image translation methods, allowing a single image representing the style of the illustrator to be applied consistently across the entire video sequence. However, selecting a style image for an illustrator is not straightforward, as the style can vary significantly between books and even within the same book, depending on the story. On the other hand, video-to-video translation approaches cannot be adopted, as there is no target video available in the original style. The only available resources are a set of unordered images in the illustrator's style, extracted from pages of the original books.

A successful video translation should meet two important criteria; capturing target style, and ensuring temporal consistency between frames.
Extending existing image-to-image translation methods ~\cite{cyclegan, pix2pix, ganilla, cartoongan, dualgan, gatsy, gatsy2, munit, unit} to independent video frames can only fulfill the first criterion. They are able to capture the target style but fail to ensure the temporal information in the video, resulting in inconsistencies between the generated frames. 
In addition to making each frame realistic, a video-to-video translation model should be capable of enhancing the temporal coherence between adjacent frames. To address these shortcomings, domain-specific methods have recently emerged~\cite{chen2017coherent, bansal2018recycle, chen2019mocycle, tip_video, wang2018vid2vid, chen2020optical, liu2021unsupervised}. 

To address the inconsistency problem temporal relationships between frames should be captured by leveraging additional information. In some approaches, optical flow is used to warp a stylized frame to estimate the next stylized frame, or it is fed to a network to generate flow in the target domain and to compare it with the source domain~\cite{chen2017coherent, gao2018reconet, ruder2018artistic, huang2017real, gao2020fast, gupta2017characterizing}. However, flow estimation is slow~\cite{chen2020optical, wang2020consistent1}; and there are no real world data sets available with ground truth flow labels. Synthetic datasets introduce noise during training due to the domain gap~\cite{sintel, dosovitskiy2015flownet}, and thus cannot be utilized efficiently.  

Alternatively, specially designed temporal predictor networks~\cite{bansal2018recycle}, knowledge distillation~\cite{chen2020optical}, or feature transformations~\cite{li2019learning} have been proposed to address the inconsistency problem. 
Although these methods achieve on-par results, there are some flaws in their design hindering their performance.  In ReCycleGan~\cite{bansal2018recycle}  input frames are fed independently diminishing the performance of the model, and optical flow is still used to train the teacher model in flow distillation~\cite{chen2020optical}.   

An alternative to optical flow warping, offering better efficiency and performance, is the feature warping approach which has been recently utilized for pose estimation, object detection, segmentation and tracking in videos~\cite{Bertasius_2020_CVPR, wu2021track, rafi2020self, bertasius2019learning}.

Inspired by the success of feature warping in other tasks, our work explores its application to unpaired video-to-video translation, addressing the challenges of complex spatio-temporal contexts.
Unlike previous methods that rely on optical flow or additional networks to ensure temporal consistency, our approach integrates a feature warping network directly into the generator. This simplifies the complex architectures for video style transfer, while also resulting in faster training and testing times. Additionally, its performance is on par with or superior to state-of-the-art methods.

In summary, our study makes two key contributions to the video style transfer problem: a new task to convert source videos of animation movies into the styles of arbitrarily selected pages from illustrations in the target domain; and a new method based on a generator network with feature warping layers, which is shown to be effective not only on task specific datasets but also on general purpose datasets.  Our method, that is ``feature {\bf W}arping for {\bf A}nimation to {\bf I}llustration video {\bf T}ranslation'', will be referred to as WAIT in short.

In the following, after reviewing the related literature~\ref{sec:rel_work}, we first describe both image-to-image and video-to-video translation methods that served as baselines, then we present our method in detail~\ref{sec:methods}. Section~\ref{sec:ill_dataset} explains the datasets used in the experiments. In Section~\ref{sec:exp}, we demonstrate the effectiveness of our method using qualitative and quantitative analysis on three challenging datasets, and we validate the design choices of WAIT through detailed ablation experiments. 

%-------------------------------------------------------------------------
\section{Related Work}
\label{sec:rel_work}
In this section, we first explain prominent GAN~\cite{gan, GAN2024111353} based image-to-image translation methods. Then, we elaborate on video-to-video translation methods.

\paragraph{\bf{Image-to-Image Translation}} Recently, several methods have been proposed for image-to-image translation~\cite{pix2pix, cartoongan, cyclegan, dualgan, ganilla, pix2pixHD, discogan, stargan, tsgan, unit, munit,10.1145/3653021,Li2024DLSGANGA}. 
CycleGAN~\cite{cyclegan} and DualGAN~\cite{dualgan} are the pioneering unpaired image-to-image translation approaches. Both utilize a cyclic framework that consists of a couple of generators and discriminators. The first couple learns a mapping from the source to the target, while the second learns a reverse mapping. MUNIT~\cite{munit} further improves unpaired image translation to multi-modal settings. First, source and target images are encoded into separate content and style latent spaces, and then they are decoded back to the image domain by merging latent vectors.

\paragraph{\bf{Video-to-Video Translation}} Although the aforementioned image-to-image translation methods produce visually appealing images for individual frames, they fail to maintain temporal coherence and are prone to flickering artifacts~\cite{ruder2016artistic}. 
To overcome these artifacts and ensure temporal consistency between generated frames, video-to-video translation methods incorporate temporal modules~\cite{gao2018reconet, chen2017coherent, lai2018learning, chen2019mocycle, gupta2017characterizing, huang2017real, ruder2016artistic, I2V-GAN2021, roma}. 

The most common approach to achieving temporal consistency is the use of optical flow. Several models utilize optical flow models to estimate the flow between the input or the generated frames (or both). Then, this flow is used to warp frames and to ensure temporal consistency between warped and stylized images~\cite{ruder2016artistic, huang2017real, chen2017coherent, gupta2017characterizing, gao2018reconet, wang2018vid2vid, lai2018learning, ruder2018artistic, wei2018video, wang2018vid2vid, wang2018fewshotvid2vid, chen2019mocycle, park2019preserving, frigo2019video, mallya2020world, gao2020fast, texler2020interactive, wang2020consistent1, wang2020consistent2, liu2021structure}. 

In its most basic form~\cite{huang2017real}, optical flow is estimated on the consecutive input frames. Then, this flow is warped with a previously stylized frame. The temporal loss is calculated between the current stylized frame and the warped stylized frame. 
In MocycleGAN~\cite{chen2019mocycle}, optical flow is estimated using FlowNet~\cite{ilg2017flownet} on both input and generated frames. Estimated flows for both sides are first translated to other domain using motion translators. These translated flows are used for warping and calculating the loss similar to~\cite{huang2017real}. 
This model requires 8 generators, 4 discriminators, 4 optical flow and 2 motion translation models forwards for just a single iteration. 

Temporal predictor networks~\cite{bansal2018recycle, liu2020unsupervised, tip_video, I2V-GAN2021, roma}, knowledge distillation~\cite{chen2020optical}, 3D convolution models~\cite{bashkirova2018unsupervised}, feature transformations \cite{li2019learning} and self-supervision~\cite{liu2021unsupervised} are also proposed to incorporate temporal relations.  ReCycleGAN~\cite{bansal2018recycle} is an extension of CycleGAN~\cite{cyclegan} with additional temporal predictors that aims to predict the next frame given two previous frames. This predicted frame is also translated into target domain and an additional loss is calculated between the predicted and the real frames.

In addition to natural grouping of the methods based on the supervision they have (i.e. supervised/paired or unsupervised/unpaired), current video-to-video translation methods could also be grouped into two categories based on the type of the target dataset (note that the source dataset is video sequences). The first category assumes that the target dataset contains a single style image. These methods~\cite{chen2017coherent, ruder2016artistic, huang2017real} usually utilize feed forward Convolutional Neural Networks (CNN) to perform translation. They extend image based Neural Style Transfer (NST) for videos with the addition of temporal components. On the other hand, the second group considers the target domain to contain ordered sequence of frames. In~\cite{wang2018vid2vid, bansal2018recycle} Generative Adversarial Networks (GANs) are utilized. Similar to previous category, image-to-image translation baselines are usually extended with temporal networks and loss functions.

%-------------------------------------------------------------------------
\section{Method}
\label{sec:methods}

\begin{figure*}
    \centering % Not needed
    \begin{subfigure}[b]{0.24\linewidth}
        \includegraphics[width=\textwidth]{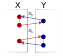}
        \caption{CycleGAN~\cite{cyclegan}}
        \label{fig:6MB_BFS}
    \end{subfigure}
    % \hfill
    \begin{subfigure}[b]{0.24\linewidth}
        \includegraphics[width=\textwidth]{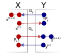}
        \caption{ReCycleGAN~\cite{bansal2018recycle}}
        \label{fig:25MB_bfs}
    \end{subfigure}
    \begin{subfigure}[b]{0.24\linewidth}
        \includegraphics[width=\textwidth]{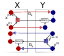}
        \caption{OpticalFlowWarp~\cite{lai2018learning}}
        \label{fig:6MB_mm}
    \end{subfigure}
    % \hfill
    \begin{subfigure}[b]{0.24\linewidth}
        \includegraphics[width=\textwidth]{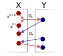}
        \caption{WAIT}
        \label{fig:25MB_mm}
    \end{subfigure}
    \caption{High level comparisons of baseline models and the proposed method WAIT. (a) CycleGAN has two generator networks: $G_X$ (red) and $G_Y$ (blue). These networks successfully capture the target styles but fails to ensure temporal coherency. (b) ReCycleGAN, has two additional temporal predictor networks, $P_X$ and $P_Y$, on top of CycleGAN to fulfill the missing feature, i.e. temporal relations. $P_X$ and $P_Y$ takes the two preceding frames, $X_{t}$ and $X_{t-1}$, and predicts the subsequent frame $X_{t+1}$. (c) OpticalFlowWarping uses \textit{Optical Flow} to capture temporal relations between generated frames. Rectangle boxes correspond to an pre-trained optical flow prediction network. Predicted flow values are used to warp with a previously generated frame to get stylized current frame.
    (d) Our method WAIT has only two networks to capture both the target style and the temporal coherency. It does not have any external network components and its design is as simple as CycleGAN. Temporal information is incorporated through feature warping layers inside the generator network. For all models, dashed lines display the loss calculations specific to the model.}
    \label{fig:all-models}
\end{figure*}

Our goal is to learn a mapping $G_X: X \rightarrow Y $ where the source domain $X$ contains \textit{ordered} sequence of frames taken from a cartoon animation movie, and the target domain $Y$ contains \textit{unordered} set of images taken from illustration books. This task differs from current video-to-video translation tasks where target domain $Y$ contains either \textit{ordered} sequences~\cite{bansal2018recycle} or a single image~\cite{huang2017real, tip_video}. 
Since there is no ground truth available for the frames in the source domain, the problem falls into unpaired translation category. 
Before going into the details of the proposed model, first we will explain the baseline methods used.
A high-level architectural description of the proposed WAIT model and comparison with baseline architectures are presented in Figure~\ref{fig:all-models}.

\subsection{Baselines}
\paragraph{{\bf CycleGAN}}
Even though CycleGAN~\cite{cyclegan} does not have a temporal component, it is a strong baseline for our task.  
CycleGAN tries to learn mappings from domain $X$ to $Y$ ($G_X$) and $Y$ to $X$ ($G_Y$) by minimizing the cyclic consistency losses. Cyclic loss for domain $X$ is given as;

\begin{equation*}
 \mathcal{L}_{c}(G_X, G_Y) = \sum_{i} \| x_i - G_Y(G_X(x_i))\|
\end{equation*}
In addition to two generators, there are two discriminators ($D_X$ and $D_Y$) that try to distinguish whether the generated images are real or fake. Adversarial loss for domain $Y$ is given as;

\begin{equation*}
\begin{aligned}
 \label{eq:adv_loss}
  \min_{G_Y} \max_{D_X} \mathcal{L}_{d}(G_Y, D_X) =  {} &  
  \sum_{j}log(D_X(y_j)) + \sum_{i}log(1 - D_X(G_Y(x_j))
  \end{aligned}
\end{equation*}
The loss function for CycleGAN is the sum of the two losses for both directions. Since CycleGAN takes a single frame and generates a single image, it only captures spatial information and ignores temporal information between frames. In~\cite{bashkirova2018unsupervised}, data is fed sequentially instead of arbitrarily and marginally better results are obtained. However, this approach still does not have any temporal component.

In order to capture the temporal information, we made simple modifications to CycleGAN. First of all, for source domain we use two frames, a reference frame ($x_t$) at time $\textit{t}$ and a nearby auxiliary frame ($x_{t+\delta}$) at time $\textit{t} + \delta$ where $\delta$ is randomly picked from the set $\{-2,...,+2\}$. The generator $G_X$ takes these two images and outputs fake images in domain $Y$, $x_{t}^{'}$ and $x_{t+\delta}^{'}$ respectively.

\paragraph{{\bf CycleGANTemp}}
We argue that foreground and background objects in nearby frames should be stylized in a similar manner. For instance, the shirt of the rabbit in Figure~\ref{fig:sample-ill} (last row third and fourth image from left) should ideally have almost the same colors in nearby frames. In other words, difference of the input frames $x_t$ and $x_{t+\delta}$ should be similar to the difference of the generated fake images $x_{t}^{'}$ and $x_{t+\delta}^{'}$. In our simple temporal baseline model, we add an additional loss for this purpose. % (Eq.~\ref{eq:diffv1_loss}). 
This baseline model will be referred to as \textit{CycleGANTemp}.

\begin{equation*}
%  \label{eq:diffv1_loss}
 \mathcal{L}_{diff}(G_X) = \sum_{i} \| (x_i - x_{i+\delta}) - (G_X(x_i) - G_X(x_{i+\delta}))\| 
\end{equation*}

\paragraph{{\bf OpticalFlowWarp}}
To compare our approach with optical flow based methods we implemented the method in ~\cite{ruder2016artistic}, which will be referred to as \textit{OpticalFlowWarp}. We calculated optical flows offline and used these pre-computed flows during training. Similar to \textit{CycleGANTemp} model, we use two frames for the source domain, except they are selected consecutively, hence the $\delta$ is fixed to $+1$ for this model. After reference and auxiliary frames are forwarded from $G_X$, we warp $x_{t+1}^{'}$ with the optical flow $F(x_t, x_{t +1})$ to get the warped version of $x_{t}^{'}$. We expect the warped stylized frame and the reference stylized frame to be identical. We use $\ell_1$ distance between $x_t$ and warped frame as the temporal consistency loss; 
% (Eq.~\ref{eq:flow_loss}).

\begin{equation*}
%  \label{eq:flow_loss}
 \mathcal{L}_{flow}(x_t, x_{t+1}) = \sum_{i} \| (x_i^{'} - W(x_{i+1}^{'}, F(x_{t+1}, x_t))\| 
\end{equation*}

\paragraph{{\bf ReCycleGAN}}
Another baseline method we experimented with is \textit{ReCycleGAN}~\cite{bansal2018recycle} and its variant adopted to our target domain. ReCycleGAN is a strong baseline for video-to-video translation problems. It integrates a temporal prediction module to capture relations between consecutive frames. However, it assumes that there is a temporal relation  both in the source and the target domains. There is almost no temporal relation between the images (i.e. illustrations) taken from the same book. For a fair comparison of ReCycleGAN with our model, we made modifications and created another version of it. We removed all temporal losses on the target domain but kept cyclic losses. We refer to this model as \textit{ReCycleGANv2}. 

\paragraph{\bf{I2V-GAN}} ~\cite{I2V-GAN2021} is another method that uses ReCycleGAN as the baseline video translation method and integrates perceptual losses~\cite{ofio2} and contrastive training~\cite{contrastive_lr} to improve the performance. Although, I2V-GAN achieves better performance compared to ReCycleGAN on multiple benchmarks, its runtime is affected by additional networks.

\subsection{WAIT}

\begin{figure*}
\centering
\includegraphics[width=1.0\textwidth]{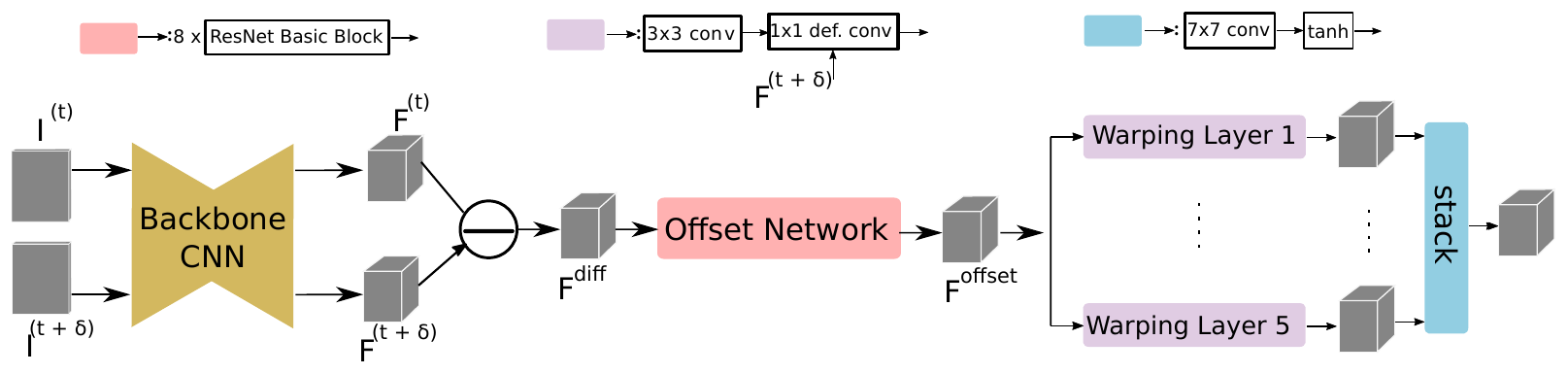}
\caption{Detailed description of the generator network of WAIT. Our model takes two images, input frame $I_t$ and auxiliary frame $I_{t+\delta}$ as inputs and forwards them through the \textit{Backbone CNN} to extract feature maps $F_t$ and $F_{t+\delta}$, respectively. Then, we calculate the difference of these two feature maps, $F_{diff}$. We forward this difference map to \textit{Offset Network} and calculate offset features $F_{offset}$. As a final stage, we warp the offset features $F_{offset}$ with the auxiliary features $F_{t+\delta}$ to create final translated frame. Warping stage contains 5 parallel layers to capture features with different resolutions.}
\label{fig:warp-model}
\end{figure*}

To capture temporal relations, our model eliminates the requirement for any auxiliary networks such as optical flow or temporal prediction networks. Instead, it warps features extracted from auxiliary frame and reference frame into a single network to generate temporally stable and stylized frames (see Figure~\ref{fig:all-models}).
We present the detailed description of the generator network of WAIT in Figure~\ref{fig:warp-model}. \\

% %-----------------------Figure Model Detayli

We use the CycleGAN network without the last convolution layer as the backbone CNN to extract $H\times W \times C$ dimensional features for reference frame $I_t$ at time $\textit{t}$ and a nearby auxiliary frame ($I_{t+\delta}$) at time $\textit{t} + \delta$ where $\delta$ is randomly picked from the set $\{-2,...,+2\}$, respectively. Then, we calculate the difference of these feature maps $\Delta_{F} = I_t - I_{t+\delta}$. This difference map is fed to the \textit{Offset Network} which consists of eight convolutional blocks where each block is composed of two sequences of a $3 \times 3$ convolution layer, an Instance Norm~\cite{IN} and a ReLU layer. These blocks are used to extract offset features. After offset features are calculated, five parallel \textit{Warping Layers} are used to warp the auxiliary frame $I_{t+\delta}$ with offset features to generate a stylized version of the input frame $I_t$. A warping layer consists of a dilated convolution and a deformable convolution layer. Dilated and deformable convolution layers have $3 \times 3$ kernels. When we consider a single warping layer, the offset features are first fed to a dilated convolution filter. The outputs of dilated convolution filter is used as offset values of deformable convolution layer and the auxiliary frame is fed to this deformable layer. We use five separate warping layers to capture offset features in different resolutions with increasing dilation values from $3$ to $24$. In the final stage, outputs of each deformable layer are stacked and then a final convolution layer with $7 \times 7$ kernels is followed by Tanh activation to generate stylized image.

Only the source domain contains temporal relations. We didn't observe any improvements using warping layers for both generators in our experiments. Therefore, in our model only $G_X$ contains warping layers and $G_Y$ contains backbone CNN with final convolution layer i.e. the same network in CycleGAN~\cite{cyclegan}. 
Since WAIT model outputs a single stylized image, we use the same training strategy and loss functions as in CycleGAN.

%-------------------------------------------------------------------------
\section{Datasets}
\label{sec:ill_dataset}

In our experiments, we have chosen examples from two illustrators who have distinct styles and whose books have been adapted to the animation movies:Beatrix Potter (BP) and Axel Scheffler (AS). We have constructed two main datasets, named using the initials of the illustrators. In both datasets, the frames of the animation movies are the source domain and the illustrations of books are the target domain. BP source split contains Peter Rabbit animation movie. AS source split contains the following animation movies; The Room on the Broom, Stickman, The Gruffalo, The Gruffalo's Child, The Highway Rat, The Snail and the Whale, and Zog. Due to the large number of frames in AS movies, we selected every 30th frame of AS movies. In  BP movies, we selected every 10th frame. For the illustrations, we have used sampled pages from several different books of BP and AS, presented in ~\cite{ganilla, 10.1145/3078971.3078982}. 
Table~\ref{tab:dataset} shows number of images in train and test sets. Example images for both datasets are shown in Figure~\ref{fig:sample-ill}.

To show the generalization capability of our method to other domains, we also use publicly available Flowers dataset which is presented in~\cite{bansal2018recycle}. This dataset contains time lapse videos corresponding to blooming of different flower species. We use the Dandelion subset which contains 2067 source and 3279 target images for training, and 700 and 1100 images for testing.

%%%%%%%%%%%-----------------------Table Dataset
\begin{table}
\centering
% \resizebox{1.0\columnwidth}{!}{
\begin{tabular}{lcccc}
\hline
\multirow{2}{*}{Dataset} &   \multicolumn{2}{c}{Train Frame Cnt} &  \multicolumn{2}{c}{Test Frame Cnt} \\
& \shortstack{Source \\ (Animation)} & \shortstack{Target \\ (Illustration)}  & Source & Target \\ 
\hline
Axel Scheffler (AS) & 9854    &  551 &  1475 & 20     \\
Beatrix Potter (BP) & 5560    &  489 &  1367 & 20     \\
Flowers             & 2067    & 3270 &  700  & 1100    \\ 
% Viper             & 90031   &   90031   & 44066 & 44066   \\ 
\hline
\end{tabular}%}
\caption{Statistics of the datasets used in experiments.}
\label{tab:dataset}
\end{table}

\section{Experiments}
\label{sec:exp}

\subsection{Implementation Details}
\label{sec:implement}

We used PyTorch~\cite{pytorch} to implement our models. As in other unpaired video-to-video transfer settings, our method does not need paired frames. Two different image datasets, one for source and the other for target, are used in our model. All training images (i.e. animation frames and illustrations) are resized to $256\times256$ pixels. We train our models for 200 epoch using Adam solver with a learning rate of 0.0008 and batch size of 8. All networks were trained from scratch. We conducted all of our experiments on Nvidia Tesla A100 GPUs. Our code, pretrained models and the scripts that reproduce the datasets can be found at https://github.com/giddyyupp/wait.

%%%%%%%%%%% Figure Visual Results AS %%%%%%%%
\begin{figure*}[t]
\captionsetup[subfigure]{labelformat=empty}
\centering
\setlength\tabcolsep{1.5pt} % default value: 6pt
\resizebox{1.0\textwidth}{!}{
\begin{tabular}{cccccc}
Input & CycleGAN & OpticalFlowWarp & ReCycleGAN & ReCycleGANv2 & WAIT \\
\animategraphics[loop,autoplay,width=4cm]{5}{./figures/results_gif/AS/input5/input5-}{0}{19} &
\animategraphics[loop,autoplay,width=4cm]{5}{./figures/results_gif/AS/cyclegan5/cyclegan5-}{0}{19} &
\animategraphics[loop,autoplay,width=4cm]{5}{./figures/results_gif/AS/optflow5/optflow5-}{0}{19} & 
\animategraphics[loop,autoplay,width=4cm]{5}{./figures/results_gif/AS/recyclegan5/recyclegan5-}{0}{19} &
\animategraphics[loop,autoplay,width=4cm]{5}{./figures/results_gif/AS/recyclegann5/recyclegann5-}{0}{19} &
\animategraphics[loop,autoplay,width=4cm]{5}{./figures/results_gif/AS/warp_new5/warp_new5-}{0}{19}\\

\hline

\animategraphics[loop,autoplay,width=4cm]{5}{./figures/results_gif/AS/input4/input4-}{0}{16} &
\animategraphics[loop,autoplay,width=4cm]{5}{./figures/results_gif/AS/cyclegan4/cyclegan4-}{0}{16} &
\animategraphics[loop,autoplay,width=4cm]{5}{./figures/results_gif/AS/optflow4/optflow4-}{0}{16} & 
\animategraphics[loop,autoplay,width=4cm]{5}{./figures/results_gif/AS/recyclegan4/recyclegan4-}{0}{16} &
\animategraphics[loop,autoplay,width=4cm]{5}{./figures/results_gif/AS/recyclegann4/recyclegann4-}{0}{16} &
\animategraphics[loop,autoplay,width=4cm]{5}{./figures/results_gif/AS/warp_new4/warp_new4-}{0}{16}\\

\end{tabular}}
\caption{Visual results on AS dataset. \textbf{Top row displays a short sequence and on the bottom a long sequence from our test set.} Leftmost column contains the input videos and the following columns correspond to results of the baseline methods and WAIT. First of all, for both sequences only WAIT captures target style correctly. In terms of visual quality and temporal coherency, on the first sequence red and black patches on the mouth of the horse are visible for every baseline. On the bottom sequence, the success of WAIT is more visible. For the CycleGAN, OpticalFlowWarp and ReCycleGAN results, it is very obvious that the temporal coherency is not captured. Especially, the color of the background grass and air changes between frames. Also, there are white/black patches/holes moving across the scene for baselines. The quality of the generated frames and temporal coherency between them are very visible for the results of WAIT.  
Zoom in for details. \textbf{Each cell is a short video clip, for the best view experience consider using a compatible (e.g. Adobe Reader) PDF reader.}}
\label{fig:visual-res-as-gifs}
\end{figure*}

%%%%%%%%%%% Figure Visual Results BP %%%%%%%%
\begin{figure*}[t]
\captionsetup[subfigure]{labelformat=empty}
\centering
\setlength\tabcolsep{1.5pt} % default value: 6pt
\resizebox{1.0\textwidth}{!}{
\begin{tabular}{cccccc}
Input & CycleGAN & OpticalFlowWarp & ReCycleGAN & ReCycleGANv2 & WAIT \\
\animategraphics[loop,autoplay,width=4cm]{5}{./figures/results_gif/BP/input3/input3-}{0}{4} &
\animategraphics[loop,autoplay,width=4cm]{5}{./figures/results_gif/BP/cyclegan3/cyclegan3-}{0}{4} &
\animategraphics[loop,autoplay,width=4cm]{5}{./figures/results_gif/BP/optflow3/optflow3-}{0}{4} & 
\animategraphics[loop,autoplay,width=4cm]{5}{./figures/results_gif/BP/recyclegan3/recyclegan3-}{0}{4} &
\animategraphics[loop,autoplay,width=4cm]{5}{./figures/results_gif/BP/recyclegann3/recyclegann3-}{0}{4} &
\animategraphics[loop,autoplay,width=4cm]{5}{./figures/results_gif/BP/warp3/warp3-}{0}{4}\\

\hline

\animategraphics[loop,autoplay,width=4cm]{5}{./figures/results_gif/BP/input2/input2-}{0}{23} &
\animategraphics[loop,autoplay,width=4cm]{5}{./figures/results_gif/BP/cyclegan2/cyclegan2-}{0}{23} &
\animategraphics[loop,autoplay,width=4cm]{5}{./figures/results_gif/BP/optflow2/optflow2-}{0}{23} & 
\animategraphics[loop,autoplay,width=4cm]{5}{./figures/results_gif/BP/recyclegan2/recyclegan2-}{0}{23} &
\animategraphics[loop,autoplay,width=4cm]{5}{./figures/results_gif/BP/recyclegann2/recyclegann2-}{0}{23} &
\animategraphics[loop,autoplay,width=4cm]{5}{./figures/results_gif/BP/warp2/warp2-}{0}{23}\\

\end{tabular}}
\caption{Visual results on BP dataset. \textbf{We display a short sequence on the top row,  and a long sequence at the bottom row} from our test set. Leftmost column contains the input videos and the following columns correspond to the results of baseline methods and WAIT respectively. First of all, similar to AS results, for both sequences only WAIT captures the target style correctly. In terms of visual quality and temporal coherency, for the first sequence, colors of the leaves and the background changes abruptly for CycleGAN, OpticalFlowWarp and ReCycleGANv2 results. For ReCycleGAN results, there is a white patch on the apple for some frames.
On the bottom sequence which is a 5 second clip, the success of WAIT is easy to catch. The style is correctly captured and the temporal coherency is ensured. For the baselines, strangely CycleGAN captures the temporal coherency much better than others. The only defect in CycleGAN results is the pinkish colorization on the wall and the head of the rabbit.
For the rest, it is very obvious that the temporal coherency is not captured. The general color palette of the scene changes dramatically between frames. Also, there are white patches/holes on the pumpkin for ReCycleGANv2 results.   
Zoom in for details. \textbf{Each cell is a short video clip, for the best view experience consider using a compatible (e.g. Adobe Reader) PDF reader.}}
\label{fig:visual-res-bp-gifs}
\end{figure*}

%-------------------------------------------------------------------------
\subsection{Qualitative Analysis}
\label{sec:qualitative}

In Figure~\ref{fig:visual-res-as-gifs}, we show example sequences from the AS test set to visually compare the results of baseline methods with our method. The results for a short clip (1.5 seconds) and for a longer clip (3 seconds) are shown on the top and the bottom rows respectively. For both sequences, all of the baseline methods except OpticalFlowWarp, generate visually plausible results. 
However, the baseline methods fail to capture the style of the illustrator, and/or lack temporal coherence. Moreover, especially in the second sequence, where we display frames from a 3 seconds clip, patches arise around objects when baseline methods are used. For instance, for ReCycleGAN and ReCycleGANv2, dark patches exist on the mouse, tail of the horse or both. These visual results align with the \textit{FID} results presented in Table~\ref{tab:as_scores}. Mode collapsed OpticalFlowWarp model has the highest FID and its visual results are the worst compared to other baselines. On the contrary, WAIT captured both temporal coherence and target style at the same time. 

Results for the BP dataset are presented in Figure~\ref{fig:visual-res-bp-gifs}. Similar to AS results, we present outputs for a short clip (1 second) and  for a longer clip (5 seconds) on the top and the bottom rows respectively. Flickering effects are easily visible on the results of the baseline methods. Colors are not consistent across the consecutive frames. For instance, on the second sequence rocks on the background become pinkish starting from the first frame on CycleGAN results. For OpticalFlowWarp, ReCycleGAN and ReCycleGANv2, intensity of colors changed on the second and the third frames. On the other hand, our model WAIT produced both correctly stylized and temporally consistent results.

%-------------------------------------------------------------------------
\subsection{Quantitative Analysis}
\label{sec:quantitative}

There are two aspects for a successful video translation: visual similarity of the translated images to the target domain and the temporal stability between stylized frames. To quantitatively evaluate the quality of video translation in terms of both aspects, we use four metrics. 

%%%% Table AS comparison.
\begin{table}
\begin{center}
% \resizebox{1.0\columnwidth}{!}{
\begin{tabular}{ccccc}
 \toprule % <-- Toprule here
 \multirow{3}{*}{Method}  &  \multicolumn{4}{c}{Metrics}      \\
 \cmidrule(lr){2-5}
  & FID$\downarrow$ & KID$\downarrow$ & FWE$\downarrow$ & MSE$\downarrow$   \\
\midrule % <-- Midrule here
CycleGAN~\cite{cyclegan} & 175.25 & 0.0602    &   0.006325    &  8528.95     \\
CycleGANTemp    & \underline{169.07}  & \underline{0.0515}          & 0.011771  & 5571.62 \\
OpticalFlowWarp~\cite{lai2018learning} & $205.69^{\dagger}$ & $0.0838^{\dagger}$  & $0.001747^{\dagger}$    &  $4566.95^{\dagger}$    \\
ReCycleGAN~\cite{bansal2018recycle} & 177.59 & 0.0547   & 0.003580    & 4397.43\\
ReCyleGANv2     & 177.16    & 0.0655     & 0.003284  & \underline{4395.39} \\

I2V-GAN~\cite{I2V-GAN2021}     & 215.60    & 0.1102        & \textbf{0.000822}  & \textbf{2446.78} \\
\cmidrule(lr){1-5}
WAIT            & \textbf{164.06}  & \textbf{0.0496} & \underline{0.003245} & 4928.44 \\
 \bottomrule % <-- Bottomrule here
\end{tabular}%}
\end{center}
\caption{Results on the AS dataset. Best results for each metric are boldfaced. $^{\dagger}$ indicates that results are not reliable, since the model failed to converge. Our method achieves best score on FID and KID metrics, and competitive result on FWE metric.}
\label{tab:as_scores}
\end{table}

Fréchet Inception Distance (FID)~\cite{FID} is a widely used metric to evaluate the visual similarity between two domains. FID compares the distributions of the features of the generated and real images and calculates the distance between them. Lower FID score yields higher visual similarity. First, real and generated images are fed to a CNN model (e.g. Inception~\cite{szegedy2016rethinking} or VGG~\cite{vgg}) trained on ImageNet~\cite{deng2009imagenet} with the classifier part removed, and activations are extracted for both domains ($X_r$ and $X_f$). Then, the mean ($m_r$ and $m_f$) and covariance ($C_r$ and $C_f$) statistics are calculated from the activations of both domains separately. The final FID score is the sum of the squared distance between the mean values and \textit{trace} of the covariance matrices:

\begin{equation*}
 \label{eq:fid_score}
 FID = \|m_r - m_f\|^{2} + tr(C_r + C_f - 2\times(C_r\times C_f)^{1/2})
\end{equation*}

Kernel Inception Distance (KID)~\cite{KID} is another popular metric to measure the visual similarity of two domains which again uses Inception weights similar to FID. The distinction lies in the calculation of the distance of domains. FID uses Fréchet distance whereas KID uses Maximum Mean Discrepancy (MMD). One advantage of KID is that it gives reliable results on small number of images compared to FID. 

\begin{equation*}
\begin{matrix}
 \label{eq:kid_score}
KID = MMD(f_{real}, f_{fake})^2 \\
k(x, y) = ((1/d) * {x}^T y + coef)^{degree}
\end{matrix}
\end{equation*}

We used the polynomial kernel function $k$ to calculate the $MMD$. $f_{real}$ and $f_{fake}$ are activations coming from real and generated images, respectively, and $d$ denotes the activation size that is $2048$ for Inception~\cite{szegedy2016rethinking}.

In order to measure temporal coherency, we use flow warping error (FWE)~\cite{huang2017real, lai2018learning} and mean squared error (MSE)~\cite{tip_video} metrics.

The \textit{Flow Warping Error (FWE)} computes the average pixel-wise
Euclidean color difference between consecutive frames:

\begin{equation*}
 \label{eq:fwe_score}
FWE = \frac{1}{T - 1} \sum_{t=1}^{T-1}\sum_{i,j,c} M_{i,j} \times  \|y_t^{i,j,c} - \hat{y_t}^{i,j,c}\|_{2}^{2}
\end{equation*}
where $y_{t}$ is the stylized framed at time $t$, $\hat{y_t}$ is the warped frame at time $t$ and which is calculated as $\hat{y_t} = W({y_t}, f(x_{t-1}, x_{t}))$. $M_{i,j}$ is a non-occlusion mask at location $i,j$ indicating non-occluded regions. We follow~\cite{tip_video} and use the occlusion detection method in~\cite{ruder2018artistic} to estimate the $M$. The $W$ function refers to flow warping operation using stylized frame $y_{t}$ and optical flow $f$ calculated between input frames $x_{t-1}$ and $x_{t}$.

The \textit{Mean Squared Error (MSE)} 
% is very similar to the loss function defined in Eq.~\ref{eq:diffv1_loss}, 
measures the errors between color differences of consecutive frames in both input and generated images:

\begin{equation*}
\begin{aligned}
%  \label{eq:mse_score}
MSE = {} 
\frac{1}{T - 1} \sum_{t=1}^{T-1}\sum_{i,j,c} & \|(x_{t+1}^{i,j,c} - x_{t}^{i,j,c}) -  ({y_{t+1}}^{i,j,c} - {y_{t}}^{i,j,c})\|_{2}^{2}
\end{aligned}
\end{equation*}
where $x_t$ and $x_{t+1}$ are two consecutive frames in the source domain, and $y_t$ and $y_{t+1}$ are corresponding stylized frames.

%%%% Table BP comparison v2.
\begin{table}
\begin{center}
%\resizebox{1.0\columnwidth}{!}{
\begin{tabular}{ccccc}
\toprule % <-- Toprule here
\multirow{3}{*}{Method} &\multicolumn{4}{c}{Metrics}  \\
 \cmidrule(lr){2-5}
& FID$\downarrow$ & KID$\downarrow$ & FWE$\downarrow$ & MSE$\downarrow$   \\
\midrule % <-- Midrule here
CycleGAN~\cite{cyclegan} & 192.55 & 0.1070  & 0.007519 &  10682.98 \\
CycleGANTemp    & \underline{165.70} & 0.0812 & 0.003998  & 7197.36 \\
OpticalFlowWarp~\cite{lai2018learning} & 166.68   & 0.0788 & 0.003850 & 6737.55 \\
ReCycleGAN~\cite{bansal2018recycle} & 172.78 & \underline{0.0782}   & 0.003781    & \underline{6493.49}\\
ReCyleGANv2     & 170.01  & 0.0896 & 0.003780  & 6896.81 \\
I2V-GAN~\cite{I2V-GAN2021}     & 273.63 & 0.2022  & \textbf{0.002615}  & \textbf{4738.73} \\
\cmidrule(lr){1-5}
WAIT   & \textbf{160.09} & \textbf{0.0717}  & \underline{0.003717} & 6745.54 \\
 \bottomrule % <-- Bottomrule here
\end{tabular}%}
\end{center}
\caption{Results on the BP dataset. Best results for each metric are boldfaced. Our method achieves best score on FID and KID metrics, and competitive result on FWE metric.}
\label{tab:bp_scores}
\end{table}

%%%% Table Flowers Dandelion comparison. v2
\begin{table}
\begin{center}
%\resizebox{1.0\columnwidth}{!}{
\begin{tabular}{ccccc}
 \toprule % <-- Toprule here
\multirow{3}{*}{Method}&\multicolumn{3}{c}{Metrics}      \\
 \cmidrule(lr){2-5}
& FID$\downarrow$ & KID$\downarrow$ & FWE$\downarrow$ & MSE$\downarrow$   \\
\midrule % <-- Midrule here
CycleGAN~\cite{cyclegan} & 184.49  & \textbf{0.1660} &  \textbf{0.000246}  &  \textbf{1153.57}   \\
CycleGANTemp    & \underline{178.98}  &  0.2255    & 0.001718  & 4694.23 \\
OpticalFlowWarp~\cite{lai2018learning} & 265.94   & 0.5547 & 0.001381 & \underline{3691.79} \\
ReCycleGAN~\cite{bansal2018recycle} & 333.53   & 0.3383 & 0.002080    & 5257.71\\
ReCyleGANv2     & 329.17   &   0.3387      & 0.001366  & 4698.50 \\
\cmidrule(lr){1-5}
WAIT            & \textbf{165.55}   & \underline{0.1992} & \underline{0.001136} & 3823.15 \\
 \bottomrule % <-- Bottomrule here
\end{tabular}%}
\end{center}
\caption{Results on the Flowers dataset. Best results for each metric are boldfaced. Our method achieves best score on FID metric.}
\label{tab:flowers_scores}
\end{table}

We present FID, KID, FWE and MSE results for baseline models and our \textit{WAIT} model in Table~\ref{tab:as_scores} and Table~\ref{tab:bp_scores} for AS and BP datasets, respectively. On all datasets, our method performs significantly better than CycleGAN and ReCyleGAN, which shows the effectiveness of our method. Specifically on the AS dataset, our method improves baseline CycleGAN by $11$ points on FID. Also, it reduces FWE and MSE metrics by almost 50\% and 40\% respectively. Similarly on the BP dataset, FID score is improved by $32$ points compared to CycleGAN. Moreover, FWE and MSE metrics are also significantly improved. Compared to ReCycleGAN, both methods achieve similar performances regarding temporal consistency metrics which shows that our model captures temporal relations with a much simpler design. 
On both datasets, I2V-GAN achieved the best scores on temporal coherency metrics thanks to its temporal prediction module. 
On the other hand, our model's visual quality performance is much better than ReCycleGAN.
Also, our modified version of ReCycleGAN performs slightly better than the original, as expected. One final remark is that our simple baseline ``CycleGANTemp" model performs quite well.

In Table~\ref{tab:flowers_scores}, we present results on the Flowers dataset. This dataset contains temporal relation on both source and target domains. Our model performed the best in terms of FID scores. However, CycleGAN outperformed others on KID, FWE and MSE metrics. The main reason for that is that the Flowers dataset is a very simple dataset that contains a fixed background and the main action happens in a very small, fixed area. FWE and MSE scores are very low for every model compared to the results of the illustration dataset.

In Table~\ref{tab:model_runtimes}, we compare all the models in terms of training and testing run times. Regarding training time we measure training a single epoch using AS dataset on a single Nvidia V100 GPU. And for testing time, we measure total time to test the Flowers dataset (including measuring the metrics) on the same GPU. For optical flow warping model, we calculated the flows offline for both train and test sets and used these pre-computed flows during training and testing. Optical flow warping models generally calculate optical flows during training on the fly to be able to randomly define the stride between selected frames. For training, our modified light version of ReCycleGAN got the smallest time. WAIT sits between the baseline CycleGAN and Flow Warping model. For test time, except ReCycleGAN all models have similar numbers since they all use a single frame to generate stylized images. 

\begin{table}
\begin{center}
%\resizebox{0.7\columnwidth}{!}{
\begin{tabular}{cccc}
 \toprule % <-- Toprule here
 \multirow{3}{*}{Method}  &  \multicolumn{2}{c}{Run Time} & \multirow{3}{*}{\shortstack{Num. \\ Params (M)}}     \\
 \cmidrule(lr){2-3}
  & Train & Test   \\
\midrule % <-- Midrule here
CycleGAN~\cite{cyclegan}        & 16              & 59  &   28.28\\
CycleGANTemp                    & 29.4            & 59  &  28.28   \\
OpticalFlowWarp~\cite{lai2018learning}  & 21.4    & 59  &  190.72 \\
ReCycleGAN~\cite{bansal2018recycle}     & 19.6    & 96  & 111.94 \\
ReCyleGANv2                     & \textbf{13.8}   & 62  &  70.11 \\
I2V-GAN~\cite{I2V-GAN2021}      & 47.1            & 66  &  104.86 \\
\cmidrule(lr){1-4}
WAIT                            & 18.8   & \textbf{54}  &  28.31 \\
 \bottomrule % <-- Bottomrule here
\end{tabular}%}
\end{center}
\caption{Comparison of methods in terms of train and test run times in minutes on a single Nvidia V100 GPU.}
\label{tab:model_runtimes}
\end{table}

%-------------------------------------------------------------------------

\subsection{Ablation Experiments}
\label{sec:abl_exp}

In order to evaluate the effects of different parts of our model and time gap parameter during training in detail, we conducted three ablation experiments. Our model consists of three main blocks, Backbone CNN, Offset Network and Warping Layers. Our Backbone CNN is amongst the established architectures for style transfer and used by many previous works such as CycleGAN~\cite{cyclegan} and ReCycleGAN~\cite{bansal2018recycle}. Offset Network and Warping Layers are two critical parts of the model, so in our ablation experiments, we focused on these two parts. We performed all ablations on the BP dataset. We trained models for $200$ epochs with a batch size of $8$.

\begin{table*}
\centering
%\resizebox{0.8\columnwidth}{!}{
\begin{tabular}{lcccc}
\toprule % <-- Toprule here
Param. & \textit{FID$\downarrow$} & \textit{KID$\downarrow$} & \textit{FWE$\downarrow$} &  \textit{MSE$\downarrow$}\\
\midrule % <-- Midrule here
2  & 166.53 & 0.0903 & 0.004115 & 7205.06      \\
4  & 162.98 & 0.0792 & 0.004179 & 7408.99      \\
6  & 172.55 & 0.0956 & 0.003733  & 6861.13      \\
8 & \textbf{160.09} & \textbf{0.0717} & \textbf{0.003717} & \textbf{6745.54}  \\
10 & 169.64 & 0.0909 & 0.005029 & 7917.83   \\
\midrule % <-- Midrule here
\multicolumn{5}{c}{(a) Effect of Offset Network Depth}\\
\bottomrule % <-- Bottomrule here
% \midrule % <-- Midrule here
1   & 170.44 & 0.0953 &  \textbf{0.003684} & 6961.53 \\
2   & 173.47 & 0.0978 & 0.003980 & 7337.73 \\
3   & 184.93  & 0.1080 & 0.004189 & 7216.33 \\
4   & 167.47  & 0.0937 & 0.003710 & 7156.28 \\
5   & 160.09 & \textbf{0.0717} & 0.003717 &  \textbf{6745.54}\\
6   & \textbf{156.13} & 0.0815 &  0.005029 &  7233.27 \\
\midrule % <-- Midrule here
\multicolumn{5}{c}{(b) Effect of Warping Layer Number}\\
\bottomrule % <-- Bottomrule here
% \midrule % <-- Midrule here
1    & 174.45 & 0.0880 & 0.005270 & 8116.54\\
2    & 160.09 & 0.0717 & \textbf{0.003717} &  \textbf{6745.54}  \\
3    & 177.37 & 0.0955 & 0.004241 & 7583.23  \\
4    & 220.92 & 0.1987 & 0.018734 & 17913.86 \\
5    & \textbf{154.94} & \textbf{0.0673} & 0.005571 & 8564.26 \\
\midrule % <-- Midrule here
\multicolumn{5}{c}{(c) Varying Time Gap}\\
\bottomrule % <-- Bottomrule here
\end{tabular}%}
\caption{Ablation experiments for the WAIT model. (a) Effect of the depth of offset network layers on performance. Offset network with $8$ residual blocks has the best performance.
(b) Effect of parallel warping layers. Here, the depth of the offset network is $8$ residual blocks and the time gap is $2$. Using $5$ warping layers gave the best performance. (c) Effect of time gap. Offset Network depth is $8$ and the number of Warping layers is $5$. Using $2$ as time gap yields the best result.}
\label{table:abl_table}
\end{table*}

\textbf{Offset Network.} We conducted ablation experiments to set the depth of the Offset Network. We used depth values of $2$, $4$, $6$, $8$ and $10$. In all these experiments, we fixed the Warping layer number to $5$ and the time gap value to $2$. Results are presented in Table~\ref{table:abl_table}a. We obtained the best performance on all metrics with depth value of $8$.

\textbf{Warping Layer.} Warping layer warps auxiliary features with offset features with different resolutions. To decide how many warping layers we should integrate in our network, we perform experiments by increasing the warping layer number from $1$ to $6$. We fixed the depth of Offset Network as $8$ and time gap value to $2$ in these ablations. In Table~\ref{table:abl_table}b, we present results of these experiments. Increasing warping layer number generally reduces the FID and KID. Using $5$ parallel warping layers gave the best performance on the KID and MSE metrics and competitive results on other metrics. Although, adding more warping layers reduces the FID further, it also slows down the training and testing times. So, we choose $5$ parallel warping layers in the final model.  

\textbf{Time Gap.} Finally, in order to determine the optimum time difference between reference and auxiliary frames, we experimented with values in the range of $1$ to $5$. Using $2$ as the time gap parameter performed the best in terms of temporal metrics. The results are given in Table~\ref{table:abl_table}c.

\subsection{Discussion}
\label{sec:disc}
Our method WAIT adds only $0.03$M parameters on top of CycleGAN and improves it on both visual similarity and temporal coherence metrics. With just a small number of additional parameters WAIT achieves similar or better performance on every metric compared to current video-to-video translation methods which rely on either optical flow warping or temporal prediction modules. These models integrate external deep networks in addition to baseline GAN networks which is reflected in the number of parameters and training times as shown in Table~\ref{tab:model_runtimes}. On the contrary, WAIT removes the need of using additional networks, simplifies the model architecture and reduces the training time to the levels of image-to-image translation methods. 

I2V-GAN achieves the best temporal coherency scores thanks to its heavy temporal prediction modules. On the other hand, its number of parameters is over $100$M and train time is more than twice slower compared to our model.

\section{Conclusions}

In this paper, we propose a new method for video-to-video translation. Our method removes the requirement for additional networks (e.g. optical flow or temporal predictors) to ensure temporal coherency between translated frames. We validated our method on a new challenging task of translating animation movies to their illustration versions. Compared to current GAN based state-of-the-art methods, our method achieved the best scores on FID and KID metrics, and produced very competitive results on FWE and MSE metrics. We conducted extensive ablation experiments to analyze different parts of our model.

%-------------------------------------------------------------------------
\section{Acknowledgements}
The numerical calculations reported in this paper were fully performed at TUBITAK ULAKBIM, High Performance and Grid Computing Center (TRUBA resources). This work is supported by the Turkish Science Academy (BAGEP).

\bibliographystyle{elsarticle-num}
\bibliography{main.bib}

\begin{thebibliography}{10}
\expandafter\ifx\csname url\endcsname\relax
  \def\url#1{\texttt{#1}}\fi
\expandafter\ifx\csname urlprefix\endcsname\relax\def\urlprefix{URL }\fi
\expandafter\ifx\csname href\endcsname\relax
  \def\href#1#2{#2} \def\path#1{#1}\fi

\bibitem{cyclegan}
J.-Y. Zhu, T.~Park, P.~Isola, A.~A. Efros, Unpaired image-to-image translation
  using cycle-consistent adversarial networks, {IEEE} International Conference
  on Computer Vision (2017).

\bibitem{pix2pix}
P.~Isola, J.-Y. Zhu, T.~Zhou, A.~A. Efros, Image-to-image translation with
  conditional adversarial networks, {IEEE} Conference on Computer Vision and
  Pattern Recognition (2017).

\bibitem{ganilla}
S.~Hicsonmez, N.~Samet, E.~Akbas, P.~Duygulu, Ganilla: Generative adversarial
  networks for image to illustration translation, Image and Vision Computing
  (2020).

\bibitem{cartoongan}
Y.~Chen, Y.-K. Lai, Y.-J. Liu, Cartoongan: Generative adversarial networks for
  photo cartoonization, in: {IEEE} Conference on Computer Vision and Pattern
  Recognition, 2018.

\bibitem{dualgan}
Z.~Yi, H.~Zhang, P.~Tan, M.~Gong, Dualgan: Unsupervised dual learning for
  image-to-image translation, {IEEE} International Conference on Computer
  Vision (2017).

\bibitem{gatsy}
L.~A. Gatys, A.~S. Ecker, M.~Bethge, A neural algorithm of artistic style,
  arXiv preprint arXiv:1508.06576 (2015).

\bibitem{gatsy2}
L.~A. Gatys, A.~S. Ecker, M.~Bethge, Image style transfer using convolutional
  neural networks, {IEEE} Conference on Computer Vision and Pattern Recognition
  (2016).

\bibitem{munit}
X.~Huang, M.-Y. Liu, S.~Belongie, J.~Kautz, Multimodal unsupervised
  image-to-image translation, in: {IEEE} European Conference on Computer
  Vision, 2018, pp. 172--189.

\bibitem{unit}
M.-Y. Liu, T.~Breuel, J.~Kautz, Unsupervised image-to-image translation
  networks, in: Advances in Neural Information Processing Systems, 2017, pp.
  700--708.

\bibitem{chen2017coherent}
D.~Chen, J.~Liao, L.~Yuan, N.~Yu, G.~Hua, Coherent online video style transfer,
  in: {IEEE} International Conference on Computer Vision, 2017, pp. 1105--1114.

\bibitem{bansal2018recycle}
A.~Bansal, S.~Ma, D.~Ramanan, Y.~Sheikh, Recycle-gan: Unsupervised video
  retargeting, in: {IEEE} European Conference on Computer Vision, 2018, pp.
  119--135.

\bibitem{chen2019mocycle}
Y.~Chen, Y.~Pan, T.~Yao, X.~Tian, T.~Mei, Mocycle-gan: Unpaired video-to-video
  translation, in: {ACM} International Conference on Multimedia, 2019, pp.
  647--655.

\bibitem{tip_video}
K.~Xu, L.~Wen, G.~Li, H.~Qi, L.~Bo, Q.~Huang, Learning self-supervised
  space-time cnn for fast video style transfer, {IEEE} Transactions on Image
  Processing 30 (2021) 2501--2512.

\bibitem{wang2018vid2vid}
T.-C. Wang, M.-Y. Liu, J.-Y. Zhu, G.~Liu, A.~Tao, J.~Kautz, B.~Catanzaro,
  Video-to-video synthesis, in: Advances in Neural Information Processing
  Systems, 2018.

\bibitem{chen2020optical}
X.~Chen, Y.~Zhang, Y.~Wang, H.~Shu, C.~Xu, C.~Xu, Optical flow distillation:
  Towards efficient and stable video style transfer, in: {IEEE} European
  Conference on Computer Vision, Springer, 2020, pp. 614--630.

\bibitem{liu2021unsupervised}
K.~Liu, S.~Gu, A.~Romero, R.~Timofte, Unsupervised multimodal video-to-video
  translation via self-supervised learning, in: Winter Conference on
  Applications of Computer Vision, 2021, pp. 1030--1040.

\bibitem{gao2018reconet}
C.~Gao, D.~Gu, F.~Zhang, Y.~Yu, Reconet: Real-time coherent video style
  transfer network, in: Asian Conference on Computer Vision, Springer, 2018,
  pp. 637--653.

\bibitem{ruder2018artistic}
M.~Ruder, A.~Dosovitskiy, T.~Brox, Artistic style transfer for videos and
  spherical images, International Journal of Computer Vision (2018) 1199--1219.

\bibitem{huang2017real}
H.~Huang, H.~Wang, W.~Luo, L.~Ma, W.~Jiang, X.~Zhu, Z.~Li, W.~Liu, Real-time
  neural style transfer for videos, in: {IEEE} Conference on Computer Vision
  and Pattern Recognition, 2017, pp. 783--791.

\bibitem{gao2020fast}
W.~Gao, Y.~Li, Y.~Yin, M.-H. Yang, Fast video multi-style transfer, in: Winter
  Conference on Applications of Computer Vision, 2020, pp. 3222--3230.

\bibitem{gupta2017characterizing}
A.~Gupta, J.~Johnson, A.~Alahi, L.~Fei-Fei, Characterizing and improving
  stability in neural style transfer, in: {IEEE} International Conference on
  Computer Vision, 2017, pp. 4067--4076.

\bibitem{wang2020consistent1}
W.~Wang, J.~Xu, L.~Zhang, Y.~Wang, J.~Liu, Consistent video style transfer via
  compound regularization, in: AAAI Conference on Artificial Intelligence,
  Vol.~34, 2020, pp. 12233--12240.

\bibitem{sintel}
N.~Mayer, E.~Ilg, P.~Hausser, P.~Fischer, D.~Cremers, A.~Dosovitskiy, T.~Brox,
  A large dataset to train convolutional networks for disparity, optical flow,
  and scene flow estimation, in: {IEEE} Conference on Computer Vision and
  Pattern Recognition, 2016.

\bibitem{dosovitskiy2015flownet}
A.~Dosovitskiy, P.~Fischer, E.~Ilg, P.~Hausser, C.~Hazirbas, V.~Golkov, P.~Van
  Der~Smagt, D.~Cremers, T.~Brox, Flownet: Learning optical flow with
  convolutional networks, in: {IEEE} International Conference on Computer
  Vision, 2015, pp. 2758--2766.

\bibitem{li2019learning}
X.~Li, S.~Liu, J.~Kautz, M.-H. Yang, Learning linear transformations for fast
  image and video style transfer, in: {IEEE} Conference on Computer Vision and
  Pattern Recognition, 2019, pp. 3809--3817.

\bibitem{Bertasius_2020_CVPR}
G.~Bertasius, L.~Torresani, Classifying, segmenting, and tracking object
  instances in video with mask propagation, in: {IEEE} Conference on Computer
  Vision and Pattern Recognition, 2020.

\bibitem{wu2021track}
J.~Wu, J.~Cao, L.~Song, Y.~Wang, M.~Yang, J.~Yuan, Track to detect and segment:
  An online multi-object tracker, in: {IEEE} Conference on Computer Vision and
  Pattern Recognition, 2021, pp. 12352--12361.

\bibitem{rafi2020self}
U.~Rafi, A.~Doering, B.~Leibe, J.~Gall, Self-supervised keypoint
  correspondences for multi-person pose estimation and tracking in videos, in:
  {IEEE} European Conference on Computer Vision, Springer, 2020, pp. 36--52.

\bibitem{bertasius2019learning}
G.~Bertasius, C.~Feichtenhofer, D.~Tran, J.~Shi, L.~Torresani, Learning
  temporal pose estimation from sparsely-labeled videos, Advances in Neural
  Information Processing Systems 32 (2019).

\bibitem{gan}
I.~Goodfellow, J.~Pouget-Abadie, M.~Mirza, B.~Xu, D.~Warde-Farley, S.~Ozair,
  A.~Courville, Y.~Bengio, Generative adversarial nets, in: Advances in Neural
  Information Processing Systems, 2014, pp. 2672--2680.

\bibitem{GAN2024111353}
Spgan: Siamese projection generative adversarial networks, Knowledge-Based
  Systems 285 (2024).

\bibitem{pix2pixHD}
T.-C. Wang, M.-Y. Liu, J.-Y. Zhu, A.~Tao, J.~Kautz, B.~Catanzaro,
  High-resolution image synthesis and semantic manipulation with conditional
  gans, in: {IEEE} Conference on Computer Vision and Pattern Recognition, 2018,
  pp. 8798--8807.

\bibitem{discogan}
T.~Kim, M.~Cha, H.~Kim, J.~K. Lee, J.~Kim, Learning to discover cross-domain
  relations with generative adversarial networks, CoRR abs/1703.05192 (2017).

\bibitem{stargan}
Y.~Choi, M.~Choi, M.~Kim, J.-W. Ha, S.~Kim, J.~Choo, Stargan: Unified
  generative adversarial networks for multi-domain image-to-image translation,
  in: {IEEE} Conference on Computer Vision and Pattern Recognition, 2018, pp.
  8789--8797.

\bibitem{tsgan}
Y.~Dong, L.~Li, L.~Zheng, Tsgan: A two-stage interpretable learning method for
  image cartoonization, Neurocomputing (2024) 127864.

\bibitem{10.1145/3653021}
Y.~Gan, C.~Yang, M.~Ye, R.~Huang, D.~Ouyang, Generative adversarial networks
  with learnable auxiliary module for image synthesis, ACM Trans. Multimedia
  Comput. Commun. Appl. (2024).

\bibitem{Li2024DLSGANGA}
W.~Li, C.~Gu, J.~Chen, C.~Ma, X.~Zhang, B.~Chen, S.~Wan, Dls-gan: Generative
  adversarial nets for defect location sensitive data augmentation, IEEE
  Transactions on Automation Science and Engineering (2024).

\bibitem{ruder2016artistic}
M.~Ruder, A.~Dosovitskiy, T.~Brox, Artistic style transfer for videos, in:
  German conference on pattern recognition, Springer, 2016, pp. 26--36.

\bibitem{lai2018learning}
W.-S. Lai, J.-B. Huang, O.~Wang, E.~Shechtman, E.~Yumer, M.-H. Yang, Learning
  blind video temporal consistency, in: {IEEE} European Conference on Computer
  Vision, 2018, pp. 170--185.

\bibitem{I2V-GAN2021}
S.~Li, B.~Han, Z.~Yu, C.~H. Liu, K.~Chen, S.~Wang, I2v-gan: Unpaired
  infrared-to-visible video translation, in: {ACM} International Conference on
  Multimedia, 2021.

\bibitem{roma}
Z.~Yu, K.~Chen, S.~Li, B.~Han, C.~H. Liu, S.~Wang, Roma: Cross-domain region
  similarity matching for unpaired nighttime infrared to daytime visible video
  translation, in: {ACM} International Conference on Multimedia, 2022.

\bibitem{wei2018video}
X.~Wei, J.~Zhu, S.~Feng, H.~Su, Video-to-video translation with global temporal
  consistency, in: {ACM} International Conference on Multimedia, 2018, pp.
  18--25.

\bibitem{wang2018fewshotvid2vid}
T.-C. Wang, M.-Y. Liu, A.~Tao, G.~Liu, J.~Kautz, B.~Catanzaro, Few-shot
  video-to-video synthesis, in: Advances in Neural Information Processing
  Systems, 2019.

\bibitem{park2019preserving}
K.~Park, S.~Woo, D.~Kim, D.~Cho, I.~S. Kweon, Preserving semantic and temporal
  consistency for unpaired video-to-video translation, in: {ACM} International
  Conference on Multimedia, 2019, pp. 1248--1257.

\bibitem{frigo2019video}
O.~Frigo, N.~Sabater, J.~Delon, P.~Hellier, Video style transfer by consistent
  adaptive patch sampling, The Visual Computer 35~(3) (2019) 429--443.

\bibitem{mallya2020world}
A.~Mallya, T.-C. Wang, K.~Sapra, M.-Y. Liu, World-consistent video-to-video
  synthesis, in: {IEEE} European Conference on Computer Vision, 2020.

\bibitem{texler2020interactive}
O.~Texler, D.~Futschik, M.~Ku{\v{c}}era, O.~Jamri{\v{s}}ka,
  {\v{S}}.~Sochorov{\'a}, M.~Chai, S.~Tulyakov, D.~S{\`y}kora, Interactive
  video stylization using few-shot patch-based training, ACM Transactions on
  Graphics (TOG) 39~(4) (2020) 73--1.

\bibitem{wang2020consistent2}
W.~Wang, S.~Yang, J.~Xu, J.~Liu, Consistent video style transfer via relaxation
  and regularization, {IEEE} Transactions on Image Processing 29 (2020)
  9125--9139.

\bibitem{liu2021structure}
S.~Liu, T.~Zhu, Structure-guided arbitrary style transfer for artistic image
  and video, IEEE Transactions on Multimedia (2021).

\bibitem{ilg2017flownet}
E.~Ilg, N.~Mayer, T.~Saikia, M.~Keuper, A.~Dosovitskiy, T.~Brox, Flownet 2.0:
  Evolution of optical flow estimation with deep networks, in: {IEEE}
  Conference on Computer Vision and Pattern Recognition, 2017.

\bibitem{liu2020unsupervised}
H.~Liu, C.~Li, D.~Lei, Q.~Zhu, Unsupervised video-to-video translation with
  preservation of frame modification tendency, The Visual Computer 36~(10)
  (2020) 2105--2116.

\bibitem{bashkirova2018unsupervised}
D.~Bashkirova, B.~Usman, K.~Saenko, Unsupervised video-to-video translation,
  arXiv preprint arXiv:1806.03698 (2018).

\bibitem{ofio2}
J.~Johnson, A.~Alahi, L.~Fei-Fei, Perceptual losses for real-time style
  transfer and super-resolution, in: {IEEE} European Conference on Computer
  Vision, Springer, 2016, pp. 694--711.

\bibitem{contrastive_lr}
T.~Chen, S.~Kornblith, M.~Norouzi, G.~Hinton, A simple framework for
  contrastive learning of visual representations, in: International Conference
  on Machine Learning, 2020, pp. 1597--1607.

\bibitem{IN}
D.~Ulyanov, A.~Vedaldi, V.~S. Lempitsky, Instance normalization: The missing
  ingredient for fast stylization, CoRR abs/1607.08022 (2016).

\bibitem{10.1145/3078971.3078982}
S.~Hicsonmez, N.~Samet, F.~Sener, P.~Duygulu, Draw: Deep networks for
  recognizing styles of artists who illustrate children's books, in:
  Proceedings of the 2017 ACM on International Conference on Multimedia
  Retrieval, 2017, p. 338–346.

\bibitem{pytorch}
A.~Paszke, S.~Gross, S.~Chintala, G.~Chanan, E.~Yang, Z.~DeVito, Z.~Lin,
  A.~Desmaison, L.~Antiga, A.~Lerer, Automatic differentiation in pytorch
  (2017).

\bibitem{FID}
M.~Heusel, H.~Ramsauer, T.~Unterthiner, B.~Nessler, S.~Hochreiter, Gans trained
  by a two time-scale update rule converge to a local nash equilibrium, in:
  Advances in Neural Information Processing Systems, 2017, pp. 6626--6637.

\bibitem{szegedy2016rethinking}
C.~Szegedy, V.~Vanhoucke, S.~Ioffe, J.~Shlens, Z.~Wojna, Rethinking the
  inception architecture for computer vision, in: {IEEE} Conference on Computer
  Vision and Pattern Recognition, 2016, pp. 2818--2826.

\bibitem{vgg}
K.~Simonyan, A.~Zisserman, Very deep convolutional networks for large-scale
  image recognition (2014).
\newblock \href {http://arxiv.org/abs/1409.1556} {\path{arXiv:1409.1556}}.

\bibitem{deng2009imagenet}
J.~Deng, W.~Dong, R.~Socher, L.-J. Li, K.~Li, L.~Fei-Fei, Imagenet: A
  large-scale hierarchical image database, in: {IEEE} Conference on Computer
  Vision and Pattern Recognition, 2009, pp. 248--255.

\bibitem{KID}
M.~Bińkowski, D.~J. Sutherland, M.~Arbel, A.~Gretton, Demystifying {MMD}
  {GAN}s, in: International Conference on Learning Representations, 2018.

\end{thebibliography}

\end{document}